# Development of an Expert System for Diabetic Type-2 Diet

Ibrahim M. Ahmed
Computer Science Department
Faculty of Computer and Information Sciences
Ain Shams University, Cairo, Egypt

Abeer M. Mahmoud
Computer Science Department
Faculty of Computer and Information Sciences
Ain Shams University, Cairo, Egypt

## ABSTRACT

A successful intelligent control of patient food for treatment purpose must combines patient interesting food list and doctors efficient treatment food list. Actually, many rural communities in Sudan have extremely limited access to diabetic diet centers. People travel long distances to clinics or medical facilities, and there is a shortage of medical experts in most of these facilities. This results in slow service, and patients end up waiting long hours without receiving any attention. Hence diabetic diet expert systems can play a significant role in such cases where medical experts are not readily available. This paper presents the design and implementation of an intelligent medical expert system for diabetes diet that intended to be used in Sudan. The development of the proposed expert system went through a number of stages such problem and need identification, requirements analysis, knowledge acquisition, formalization, design and implementation. Visual prolog was used for designing the graphical user interface and the implementation of the system. The proposed expert system is a promising helpful tool that reduces the workload for physicians and provides diabetics with simple and valuable assistance.

## General Terms

Expert Systems.

## 1. INTRODUCTION

A successful intelligent control of patient food for treatment purpose must combines patient interesting food list and doctors efficient treatment food list. Diabetes is a serious, life-threatening and chronic disease. It is estimated that this figure will reach 366 million by 2030 [1], with 81% of these diabetics being in developing countries, where medical care remains severely limited. Actually, recent estimates place the diabetes population in Sudan at around one million – around 95% of whom have type 2 diabetes and patients with diabetes make up around 10% of all hospital admissions in Sudan [2]. The types of food eaten in Sudan vary according to climate, although the Sudanese diet has plenty of carbohydrate-rich items some patients believe sugar is the only source of energy, therefore on hot days people consume large amount of sugary carbonated drinks [3]. Fortunately, diabetes can be managed very effectively through healthy lifestyle choices, primarily diet and exercise. Mostly, Type 2 diabetes is strongly connected with obesity, age, and physical inactivity [1]. Comparing with type-1 diabetes, most medical resources reported that 90 to 95% of diabetic is diagnosed as type-2. Furthermore, it can affect not only adults but young people as well. Simply, in these cases the pancreas is not able to produce enough insulin to keep the blood sugar level with in normal ranges. In addition, the majority of this type diabetics do not know they are suffering from it. Over 80-90% of

Type 2 diabetes is overweight, and this in turn contributes in many diabetes symptoms. Therefore, reducing daily carbohydrates and fats intake and the commitment to a healthy diet with a simple waling keeps your glucose with in normal ranges and help dropping those extra pounds [4].

On the other hand, the development of computer technology and tools has provided a valuable assistance for Medicare. Artificial Intelligence was primarily concerned in Medicine from the very earliest moments in the modern history of computer. It is true that the medical field is a crucial and beneficial aspect of artificial intelligence [5]. An expert system is a computer program that provides expert advice as if a real person had been consulted where this advice can be decisions, recommendations or solutions [6]. A large number of expert systems are utilized in day to day operation throughout medical research where each of these systems attempts solving part or whole of a significant problem to reduce the essential need for human experts and facilitates the effort of new graduates.

An efficient tool for diagnosing and treatment diabetes is urgently needed for helping both specialist doctors and patients. Our research was motivated by the need of such an efficient tool for diabetic. The intention of our research is to provide self-monitor for patient of type 2 diabetes to get proper amount of daily calories with list of proper diet satisfy the amount of the calories. These paper presents, the development of an intelligent expert system for diabetic type-2. The paper go through the system development cycle (the problem and need identification, requirements analysis, knowledge acquisition [7], formalization, design and implementation). Visual prolog was used for designing the graphical user interface and the implementation of the system.

The rest of paper is organized as follows. Section 2 presents literature review and related work. Diabetes food guide pyramid is in section 3. Section 4 describes the proposed expert system methodology where the proposed expert system went through a number of stages such problem and need identification, requirements analysis, knowledge acquisition, formalization, design and implementation. Conclusion and future work are given in section 5.

## 2. LITERATURES REVIEW & RELATED WORK

M.Garcia et. al (2001) [8] introduced an intelligent system to help diabetes people to monitor and to control the blood glucose level. The system named ESDIABETES. In the first phase of the ESDIABETES they presented a small prototype in CLIPS 6.0, where in final phase their system was implemented in WxCLIPS due to its friendly user interface than CLIPS 6.0. Actually the system start by asking questions and the system answers. This procedure will continue until ESDIABETES has enough information to give a recommendation to maintain the glucose level in the blood within acceptable values. This tool proves advantageous for testing and training purpose, thus it can be used by medical trainees to study about diabetes.

J.Cantais1 et. al (2005) [9] introduced building an environment for Health, and Knowledge Services Support. They create a dynamic knowledge environment that focus in managing heterogeneous knowledge from different sources. Also they proposed a Food Ontology which organizes foods in 13 main categories. They used hierarchical structure for designing the ontology. They suggest the size of the portion that allowed safely for diabetic type1.

P. M. Beulah et. al (2007) [10] introduced the ability to access diabetic expert system from any part of the world. They collect, organize, and distribute relevant knowledge and service information to the individuals. The project was designed and programmed via the dot net framework. The system allows the availability to detect and give early diagnosis of three types of diabetes namely type 1, 2, gestational diabetes for both adult and children.

M.Wiley et. al (2011) [11] presented diabetes management tool that monitors and controls blood glucose (BG) levels in order to avoid serious diabetic complications. They mentioned the difficult task for physicians, to manual large volumes of blood glucose data to tailor therapy of each patient. Also they describe three emerging applications that employ AI to ease this task. Actually, their system enables:

(a) automatic problems detection in BG control (b) offering solutions to the detected problems (c) remembering the effective and/or ineffective solutions for individual patients type1 diabetes (T1D). furthermore their system might be embedded in insulin pumps or smart phones to provide low-risk advice to patients in real time. Finally, they used support vector regression (SVR) model for building the system.

W.Szajnar and G.Setlak(2011)[12] proposed a concept of building an intelligence system of support diabetes diagnostics, where they implemented start-of-art method based on artificial intelligence for constructing a tool to model and analyze knowledge acquired from various sources. The initial target of their system was to function as a medical expert diagnosing diabetes and replacing the doctor in the first phase of illness. Diagnostics the sequence of dealing with their system were as flow: (1) getting patient information and symptoms (2) competing basic medical examination in details (3) based on previous information the system find out whether the patient has diabetes and decides whether it is type1 or type2. The systems used decision tree as a model for classification.

S. Kumar and B. Bhimrao (2012)[13] developed a natural therapy system for healing diabetic, they aim to help people's health and wellness, which don't cost the earth. Their main goal was to integrate all the natural treatment information of diabetes in one place using ESTA (Expert System Shell for Text Animation) as knowledge based system. ESTA has all facilities to write the rules that will make up a knowledge base. Further, ESTA has an inference engine which can use the rules in the knowledge base to determine which advice is to be given to the user. Their system begins with Consultation asking the users to select the disease (Diabetes) for which they want different type of natural treatment solution then describes the diabetes diseases and their symptoms. After that describes the Natural Care treatment solution of diabetes disease herbal and proper nutrition.

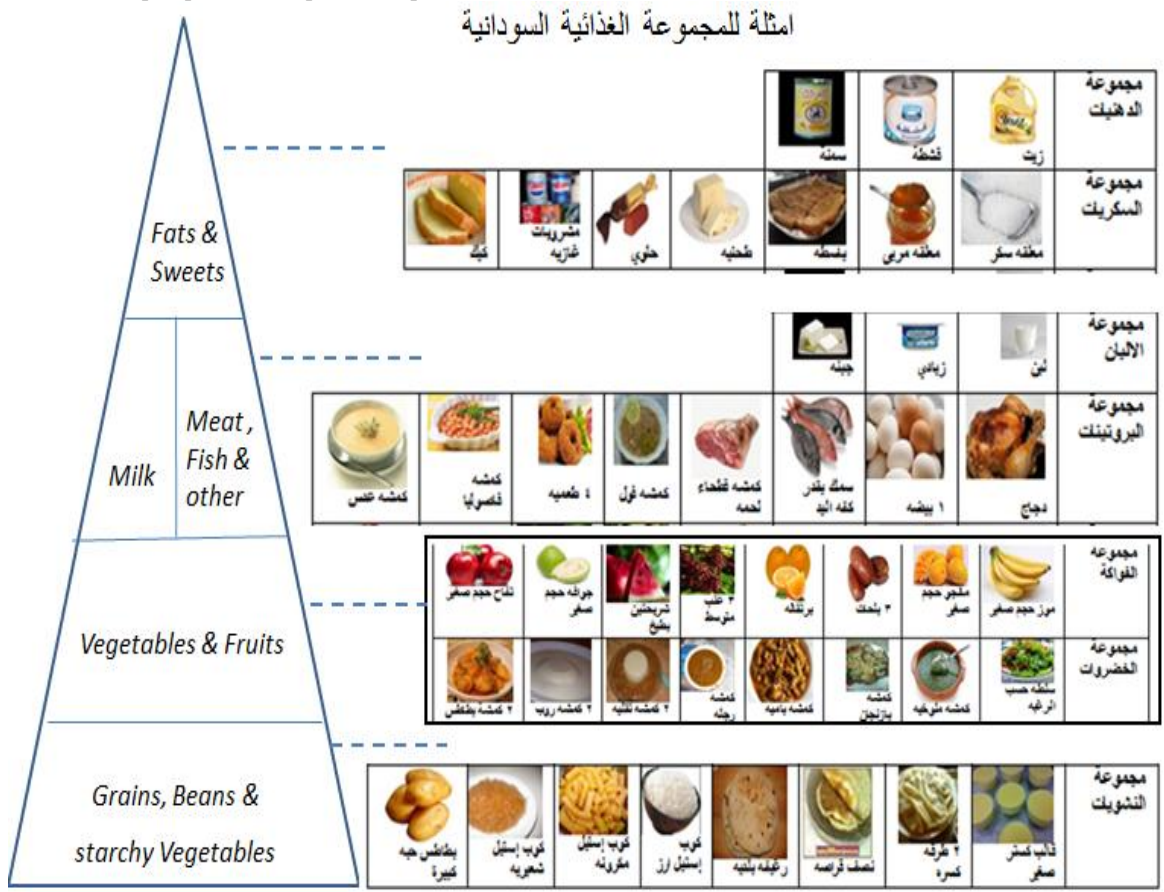

**Fig 1: Sudanese Food servings according to the Diabetes Food Guide Pyramid**

## 3. SUDANESE DIABETES FOOD GUIDE PYRAMID

The Diabetes Food Guide Pyramid is an efficient illustration of combining proper food to healthy satisfy a human and especially diabetics daily needs according to the importance of food categories. The Guide Pyramid consists of seven main croups starch, vegetables, fruits, protein, fat and sugar. Each group has similar amount of the calories. Based on the well-defined pyramid, the healthy eating must be high in nutrients, low in fat as foods that are high in carbohydrates increase blood glucose levels. [15 ]

Fig. 1 shows the Sudanese food guide pyramid based on the diabetes food guide pyramid reported in lecturers but modified according the Sudanese food culture. From the pyramid, examples of foods that increase blood glucose levels are Grains, Beans, and Starchy Vegetables group, the Fruits group, and the Milk group. Other foods that raise blood glucose are Sweets, found in the top of the Pyramid. Starchy foods, sweet foods, fruits and milk are high in carbohydrate. On the other hand Vegetables group, Meat and Others group and Fats doesn't increase blood glucose. According to the healthy pyramid diabetes patient should eat 6 to 11 servings Grains, 2 to 5 servings Group Vegetable, 2 to 4 servings Group Fruit, 2 to3 servings Group Milk, 2 to 3 servings group protein, Group sugars and oils should rarely be eaten [14].

## 4. THE PROPOSED EXPERT SYSTEM METHODOLOGY

The study was carried out using hybrid of qualitative research methodology at the military hospital in Khartoum. The main objective of the study was to develop a medical expert system for the treatment of diabetes type 2. Actually, a sample, which included: diet specialist, clinical officers, and nurses, the knowledge elicitation instruments used included: interviews, analysis of documents, observation, and questionnaires. Actually, the specific phases used in developing the proposed expert system included: (1) Problem and need identification; (2) Requirement and system analysis; (3) Knowledge acquisition; (4) Formalization or Knowledge Modeling; (5) Design or conceptualization; (6) Implementation; (7) Testing and Maintenance.

### 4.1 Problem and Need Identification

The main objective of this phase was to identify, characterize, and define the problems the system will be expected to solve. The main problems identified include: Shortage of specialist; the other medical staff in the Division needed expert knowledge and guidance, from the specialist, on treatment of diabetes; no commercially or free expert system is available in the area of diabetes; all systems available in medical fields are in English, German, or French, there isn't a single expert system available in Arabic and finally medical expert systems are rarely available in mass-distribution format.

### 4.2 Requirements Analysis

This phase involved getting to know and understand what the users needed the system to do for them and also stipulate what the system needed to function. Mainly, for user requirements, the system should automate the medical protocols to provide the patients with medical advices and basic knowledge on diabetes diet; Offer training facility on the diagnosis and treatment for diabetes. In the other hand, the system requirements focused on hardware, software, and human (end user) skills to get the fastest, most reliable and upgradeable computer system. This was categorized as: Hardware requirements (a computer); Software requirements (Operating and application); what is expected of the user to have in order to use the system (such as medical skills and knowledge and basic computer skills).

### 4.3 Knowledge Acquisition

The knowledge acquisition methods used in this study include: interview schedules, analysis of documents where many interviews were made with the diabetes and nutrition specialist at the military hospital (Dr. Iqbal and Dr.Nazik) in Sudan. Also, several visits to the Federal Ministry of Health, Department of non-communicable diseases were occurred and meetings with specialist were held to collect documents about diabetes. For instance, during this stage, the doctors were asked to answer the following: What is diabetes; how many types of diabetes; what are the diagnosis and the treatment process; what about diabetic foods? What to eat and how much; what about sugar? What medical rules or protocols guiding to proper diet; how are these medical rules or protocols used.

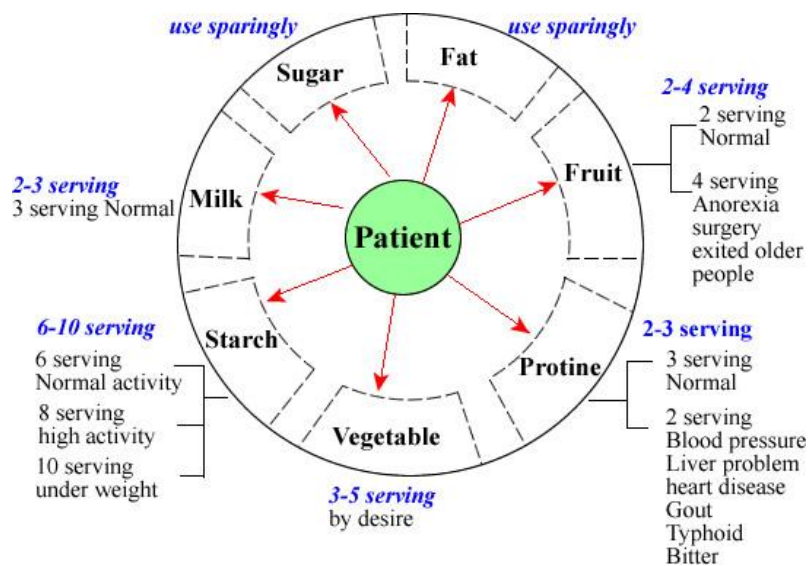

**Fig 2: Diabetics numbers of allowed servings**

```
Input : X ← body height
        Y ← body weight
        AL ← Activity Level
Output : TC ← total_Cal  // total calories

Typedefenum { " Slim", " Normal","Overweight", "Obese"} P;
Typedefenum { " Very Active", " Moderate","Little Activity"} AL;
PBMI=Calculate_BMI(X,Y); //Patient Body  Mass Index
IF PBMI<= 18.5 Then P=1;
Else { IF (PBMI>18.5 && PBMI<=25) Then P=2;
     Else {IF (PBMI>=30) Then P=3;}        }
Case P of
    1: { IF AL==1 Then TC=Y*40;
         Else TC=Y*35} Break;
    2: { IF AL==1 Then TC=Y*35;
         Else  IF AL==2 Then TC=Y*30;
             Else TC=Y*25} Break;
    3: { IF AL==1 Then TC=Y*30;
        Else  IF AL==2 Then TC=Y*25;
            Else TC=Y*20} Break;
```

**Fig 3:  Pseudo code for calculating patient total allowed calories**

**If** (anorexia=1) or (surgery=1) or (age>65)
then fruit- servings =4 else fruit- servings =2.
**If** activity="normal"
then crabs-servings=6
**Else if** activity="high"
 then crabs-servings=8
else **If** BMI<18.5
**then** crabs-servings=10.
**If** ((gout =1) or (Heart disease=1)
or (Bitter=1) or (liver problems=1) or
 (Blood pressure=1) or (Typhoid=1))
**then** protein-servings=2
**Else** protein-servings=3.
**If** ((gout =1) or (Heart disease=1) or
 (Bitter=1) or (liver problems=1) or
(Blood pressure=1) or (Typhoid=1))
 **then** milk-servings=2
 **Else** milk-servings=3.

**If** (MBI>30) then (patient is obese)
**Else** if (MBI<18.5) then (patient is slim)
**Else** (patient is normal).
**If** ((patient is slim) and (activity is very active))
**then** Total calories=weight*40.
**If** ((patient is slim) and (activity is moderate))
**then** Total calories=weight*35.
**If** ((patient is slim) and (activity is little activity))
**then** Total calories=weight*35.
**If** ((patient is normal) and (activity is very active))
**then** Total calories=weight*35.
**If** ((patient is normal) and (activity is moderate))
**then** Total calories=weight*30.
**If** ((patient is normal) and (activity is little activity))
 **then** Total calories=weight*25.

**Fig 4: rules samples of the knowledge based system of diabetics**

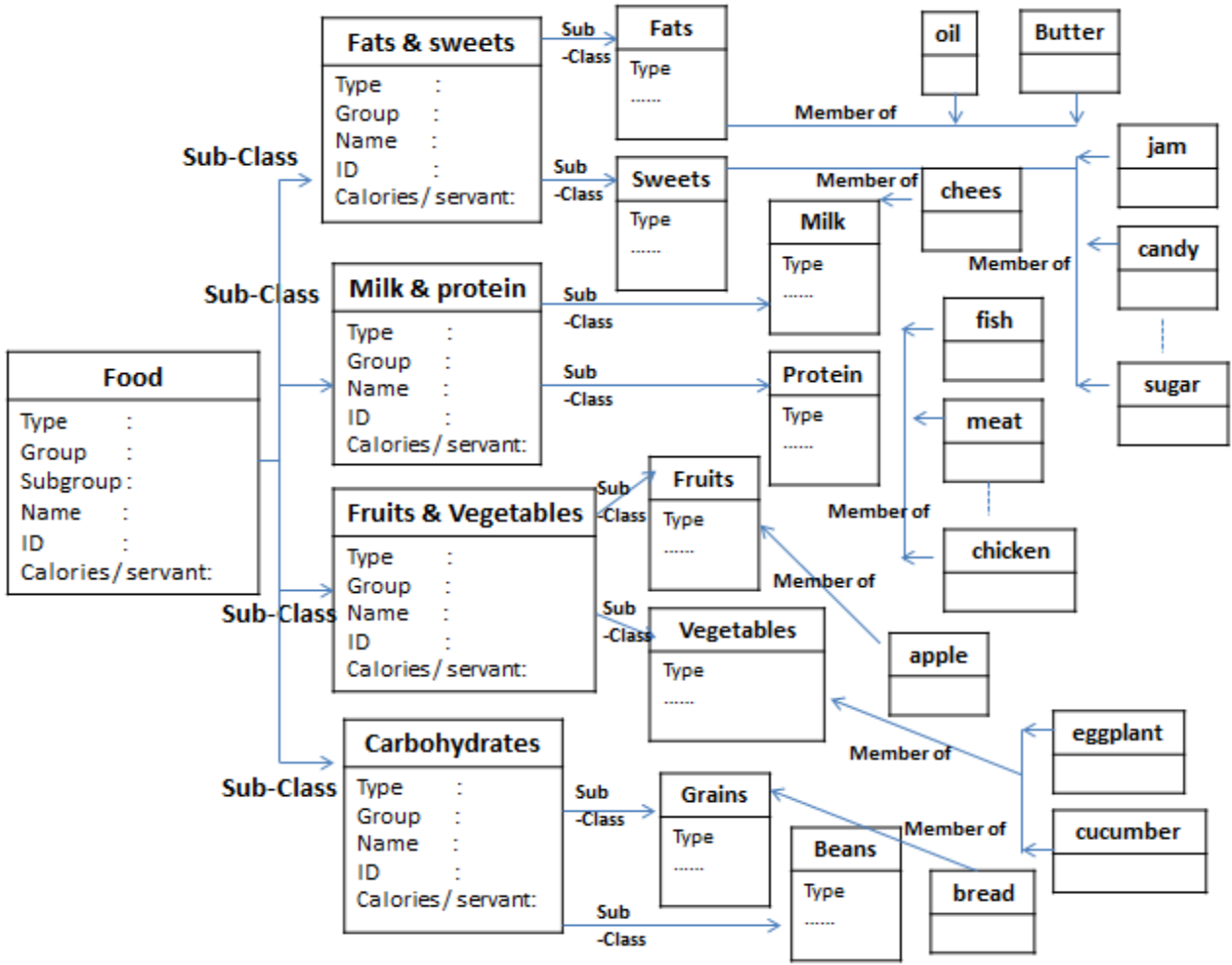

**Fig 5: Sample of diabetics food frame representation**

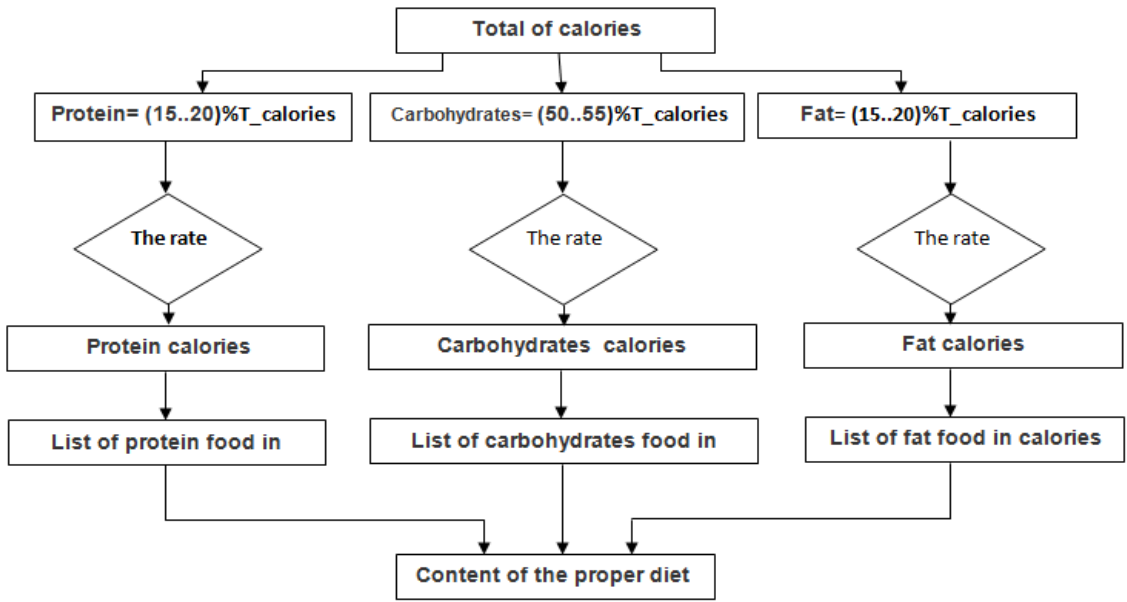

**Figure 6:  The structure of the system**

## 4.4 Formalization

The information and knowledge collected were modelled in two forms to facilitate understanding of how the system will operate and how it arrives at its conclusion: 1) for calculating food servants and calculating number of calories , a rule based representation is used.  Fig 2 and 3 show how the system determines the number of services to each patient and a Pseudo code for calculating the total calories permitted for each patient, respectively. Eq.1 shows the calculation of the Body Mass Index (BMI) [15] . Fig. 4  shows sample of the rules generated for this purpose. 2) Frame based representation is used to connect food types and subcategories of each class according to diabetics healthy food pyramid, where we find that slots provide us with more information about each Sudanese food category and subcategory and more description means better reflection of the knowledge. Fig 5 shows a sample of this frame based representation. Fig. 6 shows the proposed system  architecture.  From the figure, the system starts asking user to enter his personal information showing the patient dialogue. Based on these information the number of servings is calculated and hence appears in the next food-groups dialogue in which patient is giving permission to select   the   interested   food   list   from   system   food recommendations. Finally the system connect all gathered information and performs inferences through its knowledge engine  process to output a recommended five meals for every patient per day breakfast ,lunch ,dinner and 2 snacks .

BMI= ((Weight (kg))/〖 Height (M) 〗^2 ) ,          (1)

## 4.5 Design

The system consists of three main graphical user interface components. The first component is the Patient dialog which consist of name, gender, age, weight, height, activity type, BGL, favorite-meals and additional diseases .second is the Food groups dialog  which consist of  group name, group list ,third is  the Meals dialog which consist of  food groups , item-name, a mount, and calories. Fig. 7 and Fig.8 and fig .9 gives sample screen shots of the user interface.

## 4.6  Implementation

This phase involves the actual coding of the system (writing of the Prolog commands that run the system). The codes were developed and customized in Visual Prolog.  It runs on a Windows 7 platform, running Visual Prolog, Version 5.2 (Personal  Edition)  ,  the  prototype  consists  of  three components.

1) User interface: the interface consist of patient dialog, food group dialog and meals dialog.

2) The knowledge base: The knowledge in the knowledge base has been organized using IF-THEN rules. The general form of the rules is seen as examples in Fig.4:

 3) The inference engine: As for the inference Engine, the user provides information about the problem to be solved and the system then attempts to provide insights derived or inferred from the knowledge base by examining the facts in the knowledge base.The forward-chaining is data-driven, where data is received and hence triggers the rules where their conditions match, and move them to working memory. Rules are fired in  order of occurrence, therefore, order is of high importance. This is repeated until a conclusion is met. Our Expert System is based on user input, therefore, it uses forward chaining as its inference engine guided by the data to the goal.

 A sample of the prolog coding of the system rules can be seen in Appendix A and B.

**Fig 7: Patient dialog**

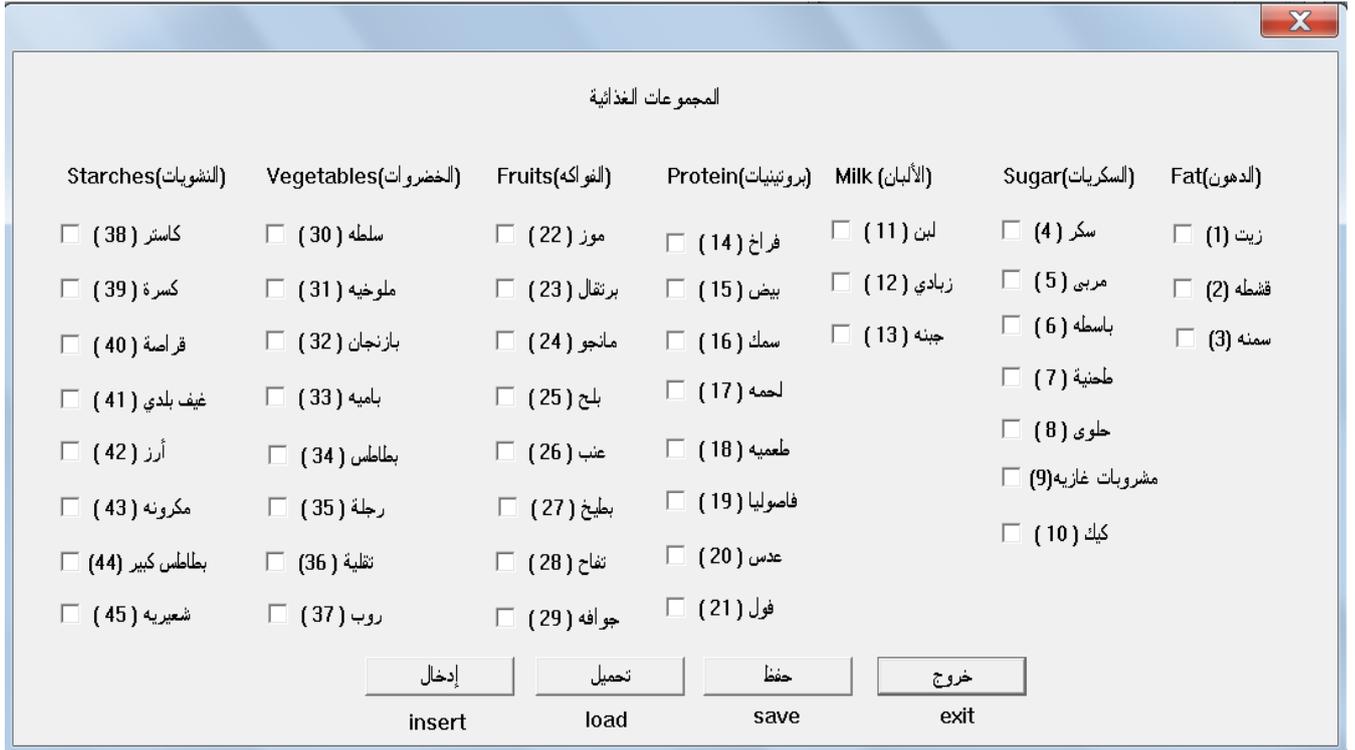

**Fig 8: Food Groups dialog**

| Where | | | | |
|---|---|---|---|---|
| In figure 6 | 1=oil | 10=pasta | ,19=bean | 28=apple | 37=roub |
| | 2=shortening | 11=milk | 20=lentils | 29=guava | 38=custer |
| | 3=synths | 12=yogurt | 21=bean | 30=salad | 39=kissra |
| | 4=sugar | 13=cheese | 22=banana | 31=Molokai | 40=gorasa |
| | 5= jam | 14=chicken | 23=orange | 32=eggplant | 41=bread |
| | 6=cake | 15=egg | 24=mango | 33=okra | 42=rice |
| | 7=tahnia | 16=fish | 25=dates | 34=potatoes | 43=pasta |
| | 8=sweet | 17=meat | 26=grapes | 35=regal | 44=potato |
| | 9= soft_drinks | 18=taamiea | 27=watermelon | 36=taglia | 45=noodles |

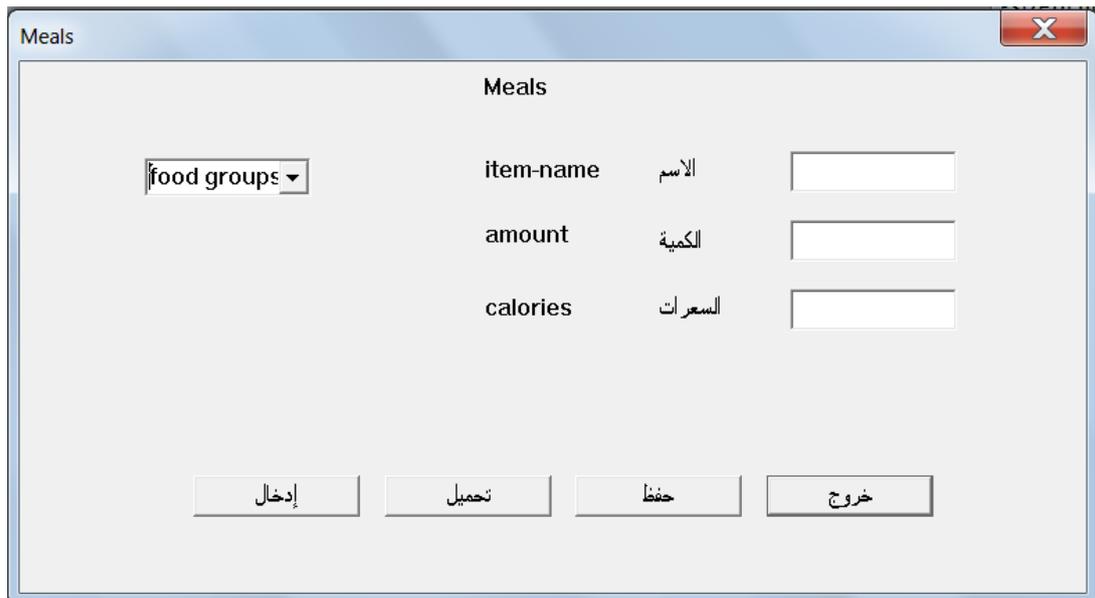

**Fig 9: Meals dialog**

## 5. CONCLUSION

This paper described the design and implementation of a medical expert system for diabetes diet that intended to be used in Sudan. The expert system provides the patients with medical advices and basic knowledge on diabetes diet. Actually, the development of the proposed expert system went through a number of stages such problem and need identification, requirements analysis, knowledge acquisition, formalization, design and implementation. Visual prolog was used for designing the graphical user interface and the implementation of the system components. The incremental development of expert systems within a rapid prototyping framework is a viable approach in the domain of diabetic assistance. It has also been important to bear in mind from the beginning of a diabetic type-2 expert system development effort that the system will eventually be used by people who are with no complex background of computer systems. Hence the graphical user interface must be a simplified as possible. In addition, the food culture of sudation must be of main attention in allowance and prevention by the recommended diet of the system. Many difficulties faced us during the requirement of various skills needed for the development of a successful diabetic expert system. These were include that more information and guidance on medication are incomplete ,thus data acquisition is expensive ,I travelled twice form Egypt to Sudan and spent 6 months collecting required information , beside the specialist is not available all the time, in addition to lack of resources no commercially or free expert system is available in the area of diabetes. The proposed expert system is a promising helpful tool that reduces the workload for physicians and provides a more comfort for diabetic patients.

In future work, the test and maintenance will investigate. Additional analysis and evaluation of the system will certainly further define the strengths and weaknesses of its approach.

## 6. REFERENCES:

[1] Audrey Mbogho et al, "Diabetes Advisor a Medical Expert System for Diabetes Management ", University of Cape Town, pp 84-87. 2005.

[2] Awad .M.Ahmed ,"Diabetes care in Sudan: emerging issues and acute needs ", Health delivery , vol. 51, no. 3, 2006.

[3] Awad.M. Ahmed and Nada Hassan Ahmed,"Diabetes mellitus in Sudan: the size of the problem and the possibilities of efficient care", Practical Diabetes Int vol.18, no.9. pp.324–327, 2001.

[4] Diabetic Diet Plan and Food Guide http://www.diabeticdietfordiabetes.com/foods.htm , 22 may, 2014

[5] Huiqing H. Yang and Sharnei Miller, "A PHP-CLIPS Based Intelligent System for Diabetic Self-Diagnosis", Department of Math & Computer Science, Virginia State University Petersburg, pp., 2006.

[6] Byoung-Ho Song, Kyoung-Woo Park and Tae Yeun Kim. "U-health Expert System with Statistical Neural Network", Advances on Information Sciences and Service Sciences. vol. 3, no.1, pp 54-61, 2011.

[7] Ibrahim M.A, et al," Knowledge Acquisition and Analysis for the Development of an Expert System for Diabetic Type-2 Diet", The Six International Conference on Intelligent Computing and Information Systems (ICICIS2013), Egypt, pp.14-16, 2013.

[8] Mario A Garcia, et.al, "Esdiabetes (An Expert System In Diabetes)", JCSC 16, pp 166-175, 2001.

[9] Jaime Cantais1, et.al, ."An example of food ontology for diabetes control", International Semantic Web Conference, 2005.

[10] P. M. Beulah Devamalar, V. Thulasi Bai, and Srivatsa S. K. "An Architecture for a Fully Automated Real-Time Web-Centric Expert System", World Academy of Science, Engineering and Technology, pp11-23, 2007.

[11] Matthew Wiley and Razvan Bunescu. "Emerging Applications for Intelligent Diabetes Management Cindy Marling", Association for the Advancement of Artificial Intelligence, pp. 2011.

[12] Wioletta SZAJNAR and Galina SETLAK. "A concept of building an intelligence system to support diabetes diagnostics", Studia Informatica, pp 260-270 , 2011.

[13] Sanjeev Kumar and Babasaheb Bhimrao. "Development of knowledge Base Expert System for Natural treatment of Diabetes disease", (IJACSA) International Journal of Advanced Computer Science and Applications, vol. 3, no. 3, pp 44-47 , 2012.

[14] Karen Halderson and Martha Archuleta "control your diabetic for life", College of Agriculture and Home Economics, NM state university, March, pp 631A1-631A4, 2013.

[15] Igbal.A and Nagwa. M,"health guide for diabetics", Sudan Federal ministry of health, 2010.

**Appendix A** : Sample Prolog Code of managing user data

```
%BEGIN patient, idc_insert _CtlInfo
  dlg_patient_eh(_Win,e_Control(idc_insert,_CtrlType,_CtrlWin,_CtlInfo),0):-!,
    NO=win_GetCtlHandle(_Win,idc_id_no),
    N=win_GetText(No),str_int(N,N1),
    Name=win_GetCtlHandle(_Win,idc_name),
    X=win_GetText(Name),
    Age= win_GetCtlHandle(_Win,idc_age),
    Y1= win_GetText(Age),str_int(Y1,Y),
     Address=win_GetCtlHandle(_Win,idc_address),
    D=win_GetText(Address),
    Wieght=win_GetCtlHandle(_Win,idc_wieght),
    W=win_GetText(Wieght),str_real(W,W1),
    Hieght= win_GetCtlHandle(_Win,idc_hieght),
    H= win_GetText(Hieght),str_real(H,H1),
    Bgl=win_GetCtlHandle(_Win,idc_bgl),
    B=win_GetText(Bgl),str_real(B,B1),
   assert(patient(N1,X,Y,D,W1,H1,B1)),
```

**Appendix B** : Sample Prolog Code of managing Food data

```
%BEGIN food_group, e_Create
  dlg_food_group_eh(_Win,e_Create(_CreationData),0):-!,
       assert(count(2)),!.
%END food_group, e_Create

%BEGIN food_group, idc_load _CtlInfo
  dlg_food_group_eh(_Win,e_Control(idc_load,_CtrlType,_CtrlWin,_CtlInfo),0):-!,
       consult("C:\\database1\\third.txt", db),
       !.
%END food_group, idc_load _CtlInfo

%BEGIN food_group, idc_save _CtlInfo
  dlg_food_group_eh(_Win,e_Control(idc_save,_CtrlType,_CtrlWin,_CtlInfo),0):-!,
       save("C:\\database1\\third.txt", db),
       !.
%END food_group, idc_save _CtlInfo
C37=win_GetCtlHandle(_Win,idc_custer),
P37= win_IsChecked(C37),
C38=win_GetCtlHandle(_Win,idc_kissra),
P38= win_IsChecked(C38),
```